\documentclass[letterpaper]{article} % DO NOT CHANGE THIS
\usepackage[preprint]{aaai2027}
\usepackage[hyphens]{url}  % DO NOT CHANGE THIS
\usepackage{graphicx} % DO NOT CHANGE THIS
\usepackage{natbib}  % DO NOT CHANGE THIS AND DO NOT ADD ANY OPTIONS TO IT
\usepackage{caption} % DO NOT CHANGE THIS AND DO NOT ADD ANY OPTIONS TO IT
\usepackage{algorithm}
\usepackage{algorithmic}
\usepackage{soul}
\usepackage{url}
\usepackage[utf8]{inputenc}
\usepackage[switch]{lineno}
\usepackage{amsthm} 

\usepackage{amsmath,amssymb,amsfonts}
\usepackage{textcomp}
\usepackage{multirow}
\usepackage{subfigure}
\usepackage{array}
\usepackage{colortbl}
\usepackage{xcolor}
\usepackage{mathrsfs}
\usepackage{newfloat}
\usepackage{listings}
\DeclareCaptionStyle{ruled}{labelfont=normalfont,labelsep=colon,strut=off} % DO NOT CHANGE THIS
\floatstyle{ruled}
\newfloat{listing}{tb}{lst}{}
\floatname{listing}{Listing}

\usepackage{booktabs}

\title{FloAff-Kitchen: Bridging Navigation and Manipulation via Canonical and Progressive Floor Affordance Learning}
\author{
    Ping Zhong,
    Manling Teng,
    Tao Wu,
    Bolei Chen\corresponding,
    Jiazhi Xia,
    Jianxin Wang \\
}
\affiliations{
    School of Computer Science and Engineering, Central South University\\
    \{ping.zhong,  manlingteng, alexanderwu2005, boleichen, xiajiazhi\}@csu.edu.cn, jxwang@mail.csu.edu.cn
    
}

\newtheorem{definition}{Definition}
\newtheorem{hypothesis}{Hypothesis}

\begin{document}

\maketitle

\begin{abstract}
Mobile manipulation requires robots to identify \textbf{Flo}or \textbf{Aff}ordance (FloAff) that maximizes downstream manipulation success rather than merely ensuring navigation feasibility. FloAff prediction is a target-conditioned local spatial reasoning problem, yet existing methods suffer from representation ambiguity caused by irrelevant spatial context and arbitrary object orientations, while entangling shared and task-specific knowledge across heterogeneous manipulation skills. To address these challenges, we propose a unified framework for FloAff prediction from egocentric multimodal perception, consisting of canonical representation learning and progressive affordance prior learning. Specifically, we introduce a \textbf{C}anonical \textbf{F}loor \textbf{A}ffordance \textbf{R}epresentation (CFAR), which learns canonical interaction geometry by preserving affordance-relevant local structure while eliminating nuisance spatial variations unrelated to robot base placement. We further propose \textbf{P}rogressive \textbf{F}loor \textbf{A}ffordance \textbf{L}earning (PFAL), which learns transferable FloAff priors from a foundation manipulation task and progressively adapts them to heterogeneous downstream manipulation skills. To facilitate systematic evaluation, we establish the first cross-scene, multi-view FloAff-Kitchen benchmark covering diverse manipulation skills, scene layouts, furniture styles, and viewpoints. Extensive experiments on three benchmark settings demonstrate that our method consistently outperforms strong baselines, while ablation studies validate the contribution of each proposed component. Project page: \url{https://csu-hero-lab.github.io/FloAff-Kitchen_Web/}.

\end{abstract}


\section{Introduction}

\begin{figure}[!t]
 \centering
 \includegraphics[width=1.0\linewidth]{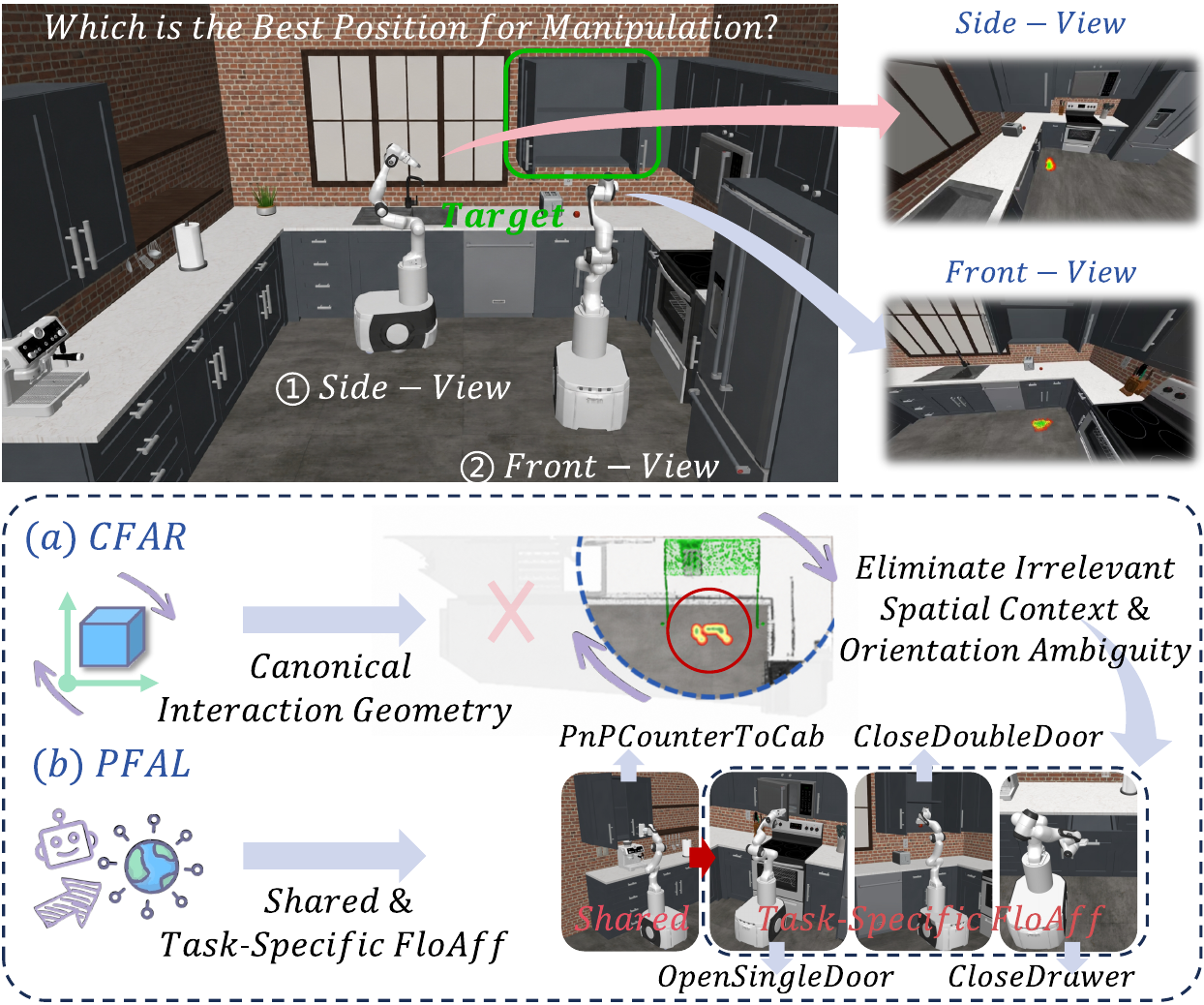}
 % \vspace{-0.5cm}
 \caption{(Top) Examples of side-view and front-view MoMa in a manipulation-rich kitchen scene. (Bottom) CFAR learns canonical interaction geometry by eliminating representation ambiguity. PFAL progressively learns transferable FloAff priors before adapting them to heterogeneous manipulation skills.}
 \label{fig1}
 % \vspace{-0.17cm}
\end{figure}

\textbf{Mo}bile \textbf{Ma}nipulation (MoMa) requires robots to determine robot base placements \cite{gu2022multi} that maximize downstream manipulation success rather than merely ensuring navigation feasibility. Consequently, \textbf{Flo}or \textbf{Aff}ordance (FloAff) prediction, \emph{i.e.}, estimating robot base placement regions that support successful manipulation from egocentric observations, has become a fundamental capability for bridging navigation and manipulation. Although recent learning-based approaches~\cite{zhang2025moma, chai2025n2m} have shown promising performance, they often learn ambiguous scene representations with irrelevant spatial context and entangle shared and task-specific knowledge across heterogeneous manipulation skills, limiting the learning of transferable FloAff priors. Following existing studies \cite{zhang2025moma}, we focus on kitchen environments, where robots perform diverse everyday manipulation skills involving objects, drawers, cabinets, and articulated appliances, as shown in Figure \ref{fig1}. However, existing benchmarks primarily evaluate front-view observations, leaving the influence of viewpoint variation on FloAff prediction largely unexplored.

We observe that FloAff is primarily determined by canonical local interaction geometry rather than the entire observed scene. Therefore, FloAff prediction is fundamentally a target-conditioned local spatial reasoning problem. However, existing methods~\cite{zhu2025navi2gaze, zhang2025moma} typically encode complete scene point clouds, introducing irrelevant spatial context. Moreover, identical interaction geometries under different object orientations produce inconsistent point cloud representations for the same affordance distribution, resulting in representation ambiguity. To address these challenges, we propose a \textbf{C}anonical \textbf{F}loor \textbf{A}ffordance \textbf{R}epresentation (CFAR), which transforms egocentric observations into a canonical representation that preserves affordance-relevant interaction geometry while eliminating nuisance variations unrelated to robot base placement. Specifically, CFAR extracts target-conditioned local geometry and aligns orientations to establish spatial correspondence, enabling affordance-equivalent scenes to be mapped into consistent representations.

Beyond representation learning, another challenge is learning transferable FloAff priors across heterogeneous manipulation skills. Although skills such as pick-and-place, door manipulation, and drawer manipulation exhibit different interaction dynamics, they share common spatial requirements for robot base placement, including reachability, collision-free execution, visibility, and manipulability. Existing multi-task learning methods \cite{zhang2025moma, chai2025n2m} optimize all skills jointly, entangling shared and task-specific affordance knowledge. We therefore propose \textbf{P}rogressive \textbf{F}loor \textbf{A}ffordance \textbf{L}earning (PFAL), which first learns transferable FloAff priors from a foundation manipulation task and then progressively adapts them to heterogeneous downstream manipulation skills.

To facilitate systematic evaluation, we further establish a cross-scene, multi-view FloAff-Kitchen benchmark based on RoboCasa \cite{robocasa2024}. We focus on kitchen environments due to their diverse manipulation requirements. Unlike existing datasets \cite{zhang2025moma}, which primarily collect front-view observations, our benchmark includes both front-view and side-view observations across diverse manipulation skills, kitchen layouts, and furniture styles, enabling systematic evaluation of viewpoint robustness and scene generalization.

We evaluate our framework on three benchmark settings built upon MoMa-Kitchen and RoboCasa. Extensive comparisons with strong baselines demonstrate that our method consistently outperforms existing methods in FloAff prediction and downstream MoMa performance, while ablation studies further validate the contribution of each component. Our main contributions are as follows:

(1) We formulate FloAff prediction as a target-conditioned local spatial reasoning problem and propose CFAR, which learns canonical interaction geometry by eliminating representation ambiguity caused by irrelevant spatial context and arbitrary object orientations.

(2) We propose PFAL, which learns transferable FloAff priors from a foundation manipulation task and progressively adapts them to heterogeneous downstream manipulation skills, effectively disentangling shared and task-specific affordance knowledge.

(3) We establish the first cross-scene, multi-view FloAff-Kitchen benchmark with explicit front-view and side-view evaluation protocols, providing a systematic testbed for studying viewpoint robustness and scene generalization. All benchmark data, evaluation protocols, and implementation code will be publicly released to facilitate future research.

\section{Related Work}

\textbf{Navigation.}
Mobile robot navigation has advanced substantially through motion planning~\cite{wu2021st}, 
semantic mapping~\cite{gervet2023navigating, hughes2022hydra, bavle2023s}, 
and learning-based policies~\cite{chen2023think, chen2025treasure}. Recent reinforcement learning, vision-language models (VLMs), 
and large language models (LLMs) further improve object-goal navigation in unseen environments by following high-level semantic instructions~\cite{kang2024hspnav, wang2026expand, yin2025unigoal}. 
However, these methods mainly optimize navigation feasibility rather than manipulation readiness. In MoMa~\cite{zhong2026spatially}, robot base placement must additionally satisfy reachability, visibility, and collision-free manipulation constraints. Consequently, heuristic navigation goals, such as stopping in front of or within a fixed distance from the target, may lead to suboptimal standing positions for downstream manipulation.

\noindent\textbf{Manipulation.}
Learning-based robotic manipulation has achieved significant progress through large-scale demonstrations, reinforcement learning, and foundation models~\cite{mandlekar2020iris, zhao2023learning, chi2025diffusion, lin2025data}. Recent vision-language-action (VLA) models further improve policy generalization by scaling model capacity and training data~\cite{kim2026cosmos, zhou2025one}. Nevertheless, successful MoMa still depends on appropriate robot base placement, as viewpoint, visibility, reachability, and surrounding geometry directly influence policy execution. Existing efforts~\cite{li2025scalable, yang2025instructvla} mainly improve manipulation policies through larger datasets or stronger models. In contrast, we investigate FloAff prediction from egocentric observations to optimize robot base placement, which is complementary to policy learning and can be integrated with different manipulation policies.

\noindent\textbf{Bridging Navigation and Manipulation.}
Recent studies bridge navigation and manipulation by explicitly reasoning about robot base placement~\cite{gu2022multi}. Early methods rely on geometric planning, including inverse reachability maps and optimization-based search~\cite{vahrenkamp2013robot, jauhri2022robot}, without considering learned manipulation policies. Subsequent approaches incorporate perception and learning into base placement prediction. Navi2Gaze~\cite{zhu2025navi2gaze} predicts navigation goals using VLMs, while MoMa-Pos~\cite{shao2025moma} optimizes articulated-object interactions through object-specific geometric modeling. MoMa-Kitchen~\cite{zhang2025moma} establishes the first benchmark for FloAff prediction, and N2M~\cite{chai2025n2m} further learns policy-aware base placement from manipulation rollouts. However, existing learning-based methods still suffer from ambiguous scene representations and limited transferability across heterogeneous manipulation skills. Our work addresses these limitations through canonical interaction geometry learning and progressive FloAff prior learning, and further establishes a cross-scene, multi-view FloAff benchmark for evaluating viewpoint robustness and scene generalization.

\section{Preliminaries}

\subsection{Problem Formulation and Overview}

Given a multimodal observation $\mathcal X=\{\mathcal I,\mathcal D,\mathcal S,o\}$, where $\mathcal I$ and $\mathcal D$ denote RGB and depth observations, respectively, $\mathcal S$ represents the robot proprioceptive state, and $o$ is the target manipulation object, our goal is to predict a FloAff map indicating effective robot base placements for downstream manipulation. The visual observations and target object are first transformed into an egocentric point cloud $\mathcal P=\Psi(\mathcal I,\mathcal D,o)$, where $\Psi$ denotes the back-projection and visual alignment process \cite{zhang2025moma}. The resulting point cloud $\mathcal P=\{\mathbf p_i\}_{i=1}^{N}$ captures the interaction geometry around the target and the spatial relationship between the robot and the target. Accordingly, FloAff prediction aims to learn a mapping $f_\theta:\{\mathcal{P},\mathcal{S}\} \rightarrow \mathcal A$. $\mathcal A\in[0,1]^{H\times W}$ denotes a FloAff map defined over the ground plane, where each element $\mathcal A(u,v)$ represents the probability that placing the robot base at location $(u,v)$ enables successful execution of the target manipulation task. Given a training set $\mathcal D=\{(\mathcal{P}_i,\mathcal{S}_i,\mathcal A_i)\}_{i=1}^{K}$, the predictor is optimized by minimizing:
\setlength\abovedisplayskip{0.1cm}
\setlength\belowdisplayskip{0.1cm}
\begin{equation} \small
\theta^{*}
=
\arg\min_{\theta}
\frac{1}{K}
\sum_{i=1}^{K}
\mathcal L_{\rm aff}
\!\left(
f_{\theta}(\mathcal{P}_i,\mathcal{S}_i),
\mathcal A_i
\right),
\label{eq:learning}
\end{equation}
where $\mathcal L_{\rm aff}$ denotes the FloAff prediction loss.

Although the formulation is straightforward, learning FloAff from raw egocentric observations remains challenging. First, affordance-relevant interaction geometry is easily affected by irrelevant spatial context and arbitrary target orientations, leading to representation ambiguity. Second, heterogeneous manipulation skills share common spatial affordance priors while exhibiting distinct task-specific interaction constraints, making it challenging to jointly learn transferable and task-specific affordance knowledge. To address these challenges, we propose CFAR to learn canonical interaction geometry and PFAL to progressively acquire transferable FloAff priors across heterogeneous manipulation skills.

\begin{figure}[!t]
 \centering
 \includegraphics[width=1.0\linewidth]{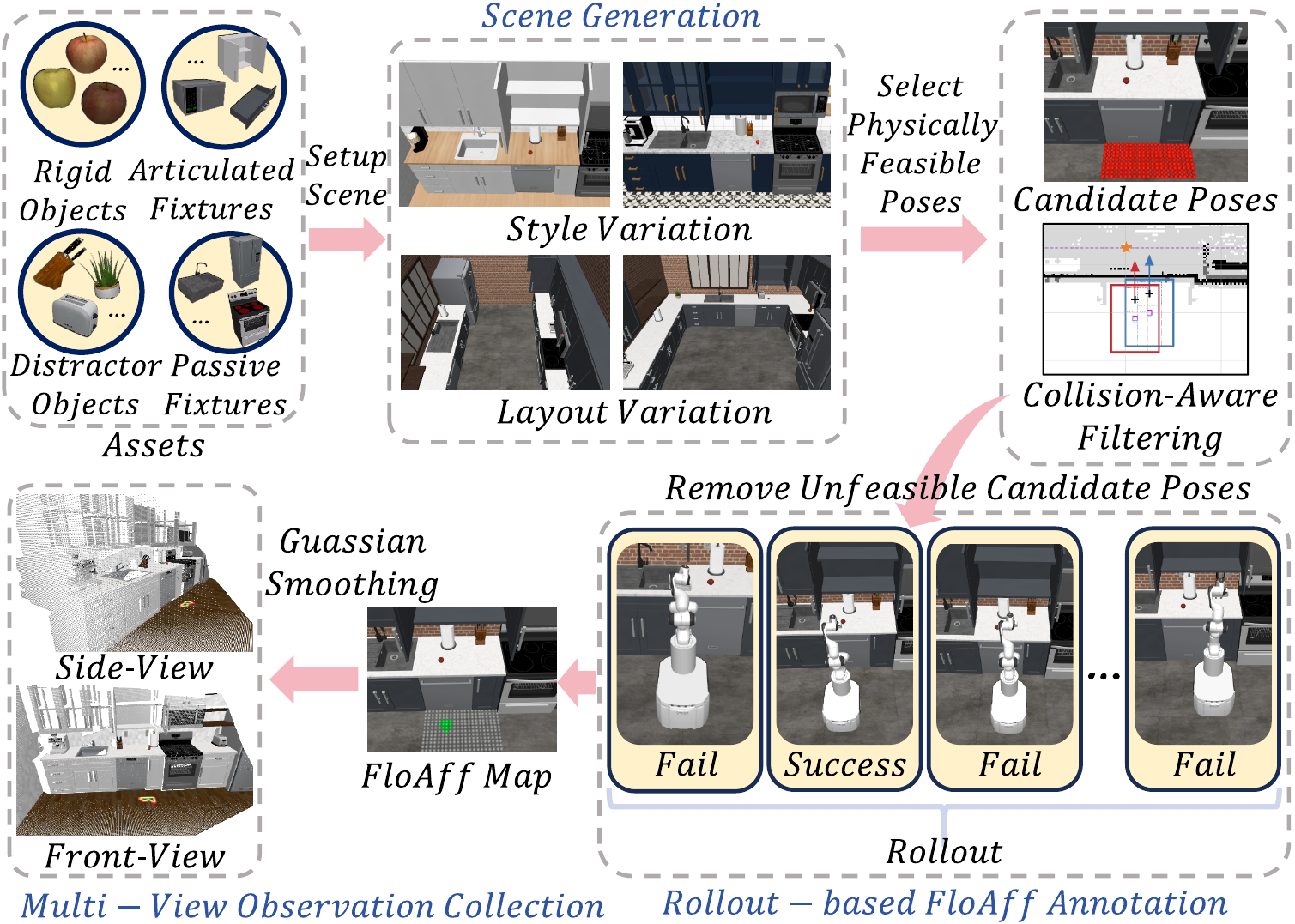}
 % \vspace{-0.5cm}
 \caption{The data collection process consists of three phases: Scene Generation, Rollout-based FloAff Annotation, and Multi-View Observation Collection. After scene generation, the candidate poses are evaluated for physical feasibility and then annotated.}
 \label{fig2}
 % \vspace{-0.18cm}
\end{figure}

% To systematically evaluate FloAff prediction, we establish a cross-scene and cross-view benchmark based on RoboCasa. 

\subsection{FloAff-Kitchen Benchmark}

Unlike existing datasets \cite{zhang2025moma} that primarily collect front-view observations for a single MoMa setting, our FloAff-Kitchen benchmark incorporates diverse manipulation skills, scene variations, and viewpoints, providing a unified testbed for evaluating viewpoint robustness and scene generalization. The following presents an overview of the data collection process (as shown in Figure \ref{fig2}). More details are provided in the supplementary materials.

\noindent\textbf{Scene Generation.}
Following existing work \cite{chai2025n2m}, we construct the benchmark in RoboCasa \cite{robocasa2024}, which provides realistic kitchen environments with configurable layouts and furniture styles. We focus on kitchen scenes because they involve diverse everyday manipulation behaviors, including object rearrangement and articulated object interaction. Specifically, our benchmark covers four representative tasks: \emph{PnPCounterToCab} (PnPC2C), \emph{OpenSingleDoor} (OpenSD), \emph{CloseDrawer} (CloseD), and \emph{CloseDoubleDoor} (CloseDD). To evaluate scene generalization, we further construct two benchmarks by independently varying kitchen layouts (\emph{Layouts} benchmark) and furniture styles (\emph{Styles} benchmark).

\noindent\textbf{Rollout-based FloAff Annotation.}
For each scene, we first generate candidate poses on the floor plane and remove physically infeasible locations through collision-aware filtering. The remaining candidate poses are evaluated by determining manipulation success using a pre-trained task-specific manipulation policy \cite{chi2025diffusion}. Let $\mathbf b_i$ denote the $i$-th candidate pose. Given the manipulation policy $\pi_\tau$ for task $\tau$, the FloAff label is defined as:
\begin{equation} \small
y_i
=
\mathbf{1}
\left[
\mathrm{Success}
(
\pi_\tau,
\mathbf b_i
)
\right],
\label{eq:rollout_label}
\end{equation}
where $y_i=1$ indicates successful task execution from pose $\mathbf b_i$, and $y_i=0$ otherwise. The rollout labels are projected onto the corresponding candidate poses to construct the \textbf{G}round-\textbf{T}ruth (GT) FloAff map $\mathcal A \in [0,1]^{H\times W}$ for supervising FloAff prediction. In practice, Gaussian smoothing is applied to encourage spatial continuity.

\noindent\textbf{Multi-View Observation Collection.}
Unlike existing datasets \cite{zhang2025moma}, our benchmark explicitly includes both front-view and side-view observations. Front-view observations provide relatively complete visibility of the manipulation target, whereas side-view observations introduce stronger occlusion and viewpoint ambiguity, especially for articulated object manipulation. Both viewpoints share identical rollout-generated FloAff labels, enabling systematic evaluation of viewpoint robustness and viewpoint-invariant FloAff prediction.

\noindent\textbf{Benchmark Statistics.}
Our benchmark contains 24,782 RGB-D observations with rollout-generated FloAff annotations, including 13,610 samples in the \emph{Styles} benchmark and 11,172 samples in the \emph{Layouts} benchmark. All train/test splits are performed at the scene level, using five training scenes and one unseen testing scene for each task.

\subsection{Canonical FloAff Representation}

Existing methods \cite{chai2025n2m,zhang2025moma} directly predict FloAff from complete scene observations, leading to two limitations: (1) FloAff is primarily determined by local interaction geometry around the manipulation target, while distant structures introduce irrelevant spatial context. (2) Affordance-equivalent scenes observed under different target orientations produce inconsistent point cloud representations, resulting in representation ambiguity. Consequently, the model must simultaneously identify affordance-relevant geometry and learn orientation invariance. To address these issues, we propose CFAR.

\begin{definition}
A CFAR is a target-conditioned scene representation that preserves affordance-relevant local geometry while being invariant to irrelevant spatial context and target orientation.
\end{definition}

Instead of directly learning $f_{\theta}(\mathcal P,\mathcal S)$, CFAR first transforms the observations into a canonical representation:
\begin{equation} \small
\hat{\mathcal A}
=
f_{\theta}\!\left(
\mathcal T(\mathcal P,\mathcal S)
\right),
\label{eq:cfar}
\end{equation}
where $\mathcal T(\cdot)$ denotes the canonical transformation. CFAR is designed to satisfy two properties: (1) \emph{Affordance Relevance}, which preserves geometry required for FloAff prediction while removing irrelevant context; and (2) \emph{Orientation Consistency}, which maps affordance-equivalent scenes into consistent representations. Specifically, $\mathcal T(\cdot)$ is achieved through the following two aspects.

\noindent\textbf{Eliminating Irrelevant Spatial Context.}
FloAff prediction primarily depends on the local interaction geometry around the manipulation target. Therefore, we construct a target-centered local representation by retaining only points within a sphere centered at the target position $\mathbf t=(x_t,y_t,z_t)$:
\begin{equation} \small
\mathcal P_{\rm local}
=
\left\{
\mathbf p_i\in\mathcal P
~
\middle|
~
\|\mathbf p_i-\mathbf t\|_2
\le r
\right\},
\label{eq:crop}
\end{equation}
where $r$ is the predefined crop radius. This transformation removes irrelevant scene context while preserving the local geometry required for FloAff prediction.

\begin{figure}[!t]
 \centering
 \includegraphics[width=1.0\linewidth]{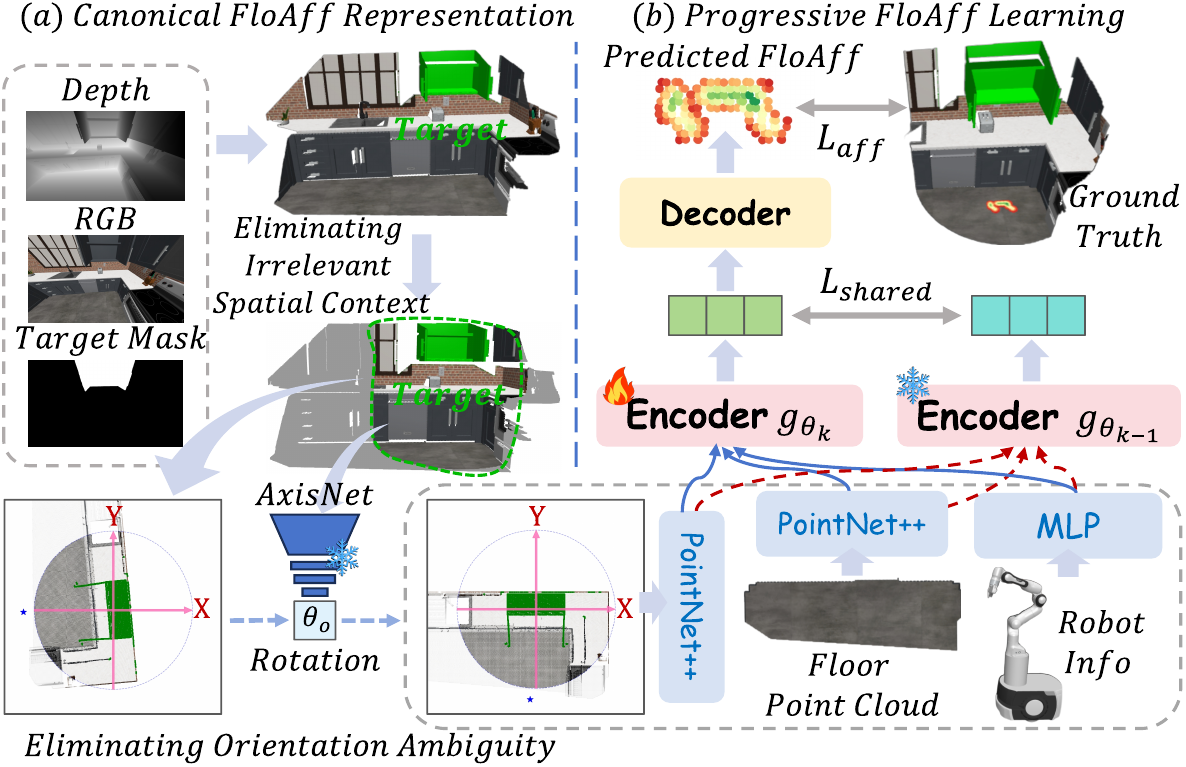}
 % \vspace{-0.5cm}
 \caption{Illustrations of (a) Canonical FloAff Representation and (b) Progressive FloAff Learning.}
 \label{fig3}
 % \vspace{-0.17cm}
\end{figure}

\noindent\textbf{Eliminating Orientation Ambiguity.}
Although local geometry is extracted, the same affordance pattern may correspond to different point cloud representations because the manipulation target can appear with arbitrary orientations. Consequently, geometrically equivalent interactions may not be spatially aligned. To establish a unified spatial correspondence, we rotate each local point cloud according to the target orientation:
% \begin{equation} \small
% \hat{\mathcal P}
% =
% \mathbf R(\pi-\theta_o)
% \mathcal P_{\rm local},
% \label{eq:rotation}
% \end{equation}
\begin{equation}
\small
\hat{\mathcal P}
=
\left\{
\mathbf R(\pi-\theta_o)
\left(\mathbf p_i-\mathbf t\right)
\;\middle|\;
\mathbf p_i \in \mathcal P_{\rm local}
\right\},
\label{eq:rotation}
\end{equation}
where $\mathbf R(\cdot)\in SO(3)$ denotes the rotation matrix about the vertical axis. During training, the target yaw angle $\theta_o$ is obtained from simulator annotations, while during inference it is estimated by a lightweight orientation predictor, AxisNet, from the cropped point cloud and target-object feature (see the supplementary material for details). As shown in Figure \ref{fig3} (a), the transformation aligns the front-facing direction of every target with the negative $y$-axis. This enables affordance-equivalent scenes to share a unified spatial representation, allowing the network to learn interaction geometry under a canonical coordinate system rather than orientation-dependent scene configurations.

\subsection{Progressive FloAff Learning}

Although heterogeneous MoMa tasks exhibit distinct interaction dynamics, they share common spatial requirements for robot base placement \cite{gu2022multi}, including reachability, collision-free execution, visibility, and manipulability, which we collectively refer to as shared FloAff priors. Meanwhile, different manipulation skills introduce task-specific affordance constraints, such as hinge directions for articulated doors and sliding directions for drawers. Existing multi-task learning methods \cite{zhang2025moma} jointly optimize multiple tasks, entangling transferable spatial reasoning with task-specific interaction patterns and limiting generalization across heterogeneous manipulation skills. We therefore introduce the following hypothesis.

\begin{hypothesis}
\label{hyp:latent_rep}
Across heterogeneous MoMa skills, FloAff prediction can benefit from a transferable latent representation that captures task-independent spatial reasoning for robot base placement, while task-specific affordance knowledge can be progressively learned through representation adaptation.
\end{hypothesis}

Based on this hypothesis, we propose PFAL and conceptually decompose the latent FloAff representation as
$
\mathbf z=\Phi(\mathbf z_s,\mathbf z_t),
$
where $\mathbf z_s$ encodes shared FloAff priors and $\mathbf z_t$ captures task-specific interaction constraints. Instead of explicitly enforcing this decomposition, PFAL progressively approximates it through a curriculum of increasing FloAff complexity:
$
\theta_{\rm pnp}
\rightarrow
\theta_{\rm door,drawer},
$
where transferable spatial reasoning is first established and then adapted to hinge- and sliding-dependent FloAff patterns.

\noindent\textbf{Shared FloAff Priors Learning.}
Pick-and-place mainly relies on generic spatial reasoning with relatively weak task-specific constraints. Therefore, we use it as the foundation task for learning shared FloAff priors. Let $\mathcal D_{\rm pnp}$ denote the pick-and-place dataset. The FloAff predictor is optimized by:
\begin{equation} \small
\theta_{\rm pnp}
=
\arg\min_{\theta}
\frac{1}{|\mathcal D_{\rm pnp}|}
\sum_{(\hat{\mathcal P}, \mathcal S, \mathcal A)\in\mathcal D_{\rm pnp}}
\mathcal L_{\rm aff}
\!\left(
f_\theta(\hat{\mathcal P},\mathcal S),
\mathcal A
\right),
\label{eq:base}
\end{equation}
where $\hat{\mathcal P}$ denotes the CFAR representation and $\mathcal A$ denotes the GT FloAff map. The learned parameters $\theta_{\rm pnp}$ are used to initialize subsequent adaptation stage. The FloAff prediction loss is defined as
$
\mathcal L_{\rm aff}
=
\mathcal L_{\rm mse}
+
\beta
\mathcal L_{\rm sdl},
$
where $\beta$ balances the two objectives. The \textbf{M}ean \textbf{S}quared \textbf{E}rror (MSE) loss supervises the FloAff probability at each floor location:
\begin{equation} \small
\mathcal L_{\rm mse}
=
\frac{1}{HW}
\sum_{u=1}^{H}
\sum_{v=1}^{W}
\left(
\hat{\mathcal A}(u,v)
-
\mathcal A(u,v)
\right)^2,
\label{eq:mse}
\end{equation}
where $\hat{\mathcal A}$ and $\mathcal A$ denote the predicted and GT FloAff maps, respectively. Since MSE alone may not preserve the spatial continuity of feasible robot base placement regions, we employ the \textbf{S}oft \textbf{D}ice \textbf{L}oss (SDL) \cite{milletari2016v}:
\begin{equation} \small
\mathcal L_{\rm sdl}
=
1-
\frac{
2\sum_{u,v}
\hat{\mathcal A}(u,v)\mathcal A(u,v)+\epsilon
}{
\sum_{u,v}\hat{\mathcal A}(u,v)^2
+
\sum_{u,v}\mathcal A(u,v)^2
+\epsilon
},
\label{eq:sdl}
\end{equation}
where $\epsilon$ is a small constant. SDL encourages overlap between the predicted and GT FloAff regions.

\noindent\textbf{Progressive FloAff Priors Learning.}
Starting from $\theta_{\rm pnp}$, PFAL progressively adapts the model by jointly optimizing the two door-related tasks, \emph{OpenSingleDoor} and \emph{CloseDoubleDoor}, which share hinge-dependent interaction geometry, together with \emph{CloseDrawer}, whose affordance depends on sliding interaction geometry. Such adaptation may shift the representation toward task-specific patterns. Therefore, as shown in Figure \ref{fig3} (b), we introduce a feature distillation objective to preserve shared FloAff priors. Let $g_{\theta_{k-1}}(\cdot)$ and $g_{\theta_k}(\cdot)$ denote the previous-stage and current feature extractors, respectively. The distillation loss is:
\begin{equation} \small
\mathcal L_{\rm shared}
=
\frac{1}{|\mathcal D_k|}
\sum_{\hat{\mathcal P}\in\mathcal D_k}
\left\|
g_{\theta_{k-1}}(\hat{\mathcal P})
-
g_{\theta_k}(\hat{\mathcal P})
\right\|_2^2,
\label{eq:shared}
\end{equation}
which encourages the preservation of transferable FloAff priors during progressive adaptation. The overall objective is:
\begin{equation} \small
\mathcal L_{pro}
=
\mathcal L_{\rm aff}
+
\lambda
\mathcal L_{\rm shared},
\label{eq:total}
\end{equation}
where $\lambda$ controls the trade-off between task-specific adaptation and shared prior preservation. At each stage, the model is optimized as:
\begin{equation} \small
\theta_k
=
\arg\min_{\theta}
\mathcal L_{pro},
\qquad
\theta
\leftarrow
\theta_{k-1},
\label{eq:adaptation}
\end{equation}
where the parameters are initialized from the previous stage.

PFAL progressively adapts the model to increasingly specialized FloAff patterns while preserving transferable FloAff priors, resulting in improved generalization across heterogeneous manipulation skills.

\section{Experiments}

\subsection{Experimental Setup}

\noindent\textbf{Dataset.}
We evaluate our method on the public MoMa-Kitchen benchmark~\cite{zhang2025moma} and two proposed FloAff-Kitchen benchmarks. Following the original split of MoMa-Kitchen, we use 456 scenes for training and 113 unseen scenes for testing. Each sample contains egocentric RGB-D observations, robot states, target masks, and dense FloAff annotations. The proposed FloAff-Kitchen benchmarks further evaluate FloAff prediction under scene and viewpoint variations. Specifically, they include two settings: (1) \emph{Styles}, which varies furniture styles and textures while preserving layouts, and (2) \emph{Layouts}, which varies kitchen layouts. Each setting covers four representative manipulation tasks (PnPC2C, OpenSD, CloseD, and CloseDD), with five training scenes and one unseen testing scene per task. Each sample contains target-centric egocentric observations collected from front-view or side-view viewpoints, paired with rollout-generated FloAff labels.

\noindent\textbf{Baselines.}
On the MoMa-Kitchen benchmark, we compare our method with three generic point cloud backbones, \textbf{PointNet++}~\cite{qi2017pointnet++}, \textbf{VoteNet}~\cite{qi2019deep}, and \textbf{H3DNet}~\cite{zhang2020h3dnet}, as well as the task-specific FloAff prediction method \textbf{NavAff}~\cite{zhang2025moma}. On the FloAff-Kitchen benchmarks, we further compare with two recent MoMa methods, \textbf{C2F-Exp}~\cite{lin2026affordance} and \textbf{N2M}~\cite{chai2025n2m}, together with NavAff.

\noindent\textbf{Metrics.}
Following existing work~\cite{zhang2025moma}, we report Root Mean Squared Error (RMSE), Logarithmic Mean Squared Error (logMSE), Pearson Correlation Coefficient (PCC), and Cosine Similarity (SIM) on MoMa-Kitchen to evaluate FloAff prediction quality. On the FloAff-Kitchen benchmarks, we additionally report affordance-grounded manipulation Success Rate (SR) to evaluate the downstream MoMa performance enabled by FloAff prediction.

\begin{figure*}[!t]
 \centering
 \includegraphics[width=1.0\linewidth]{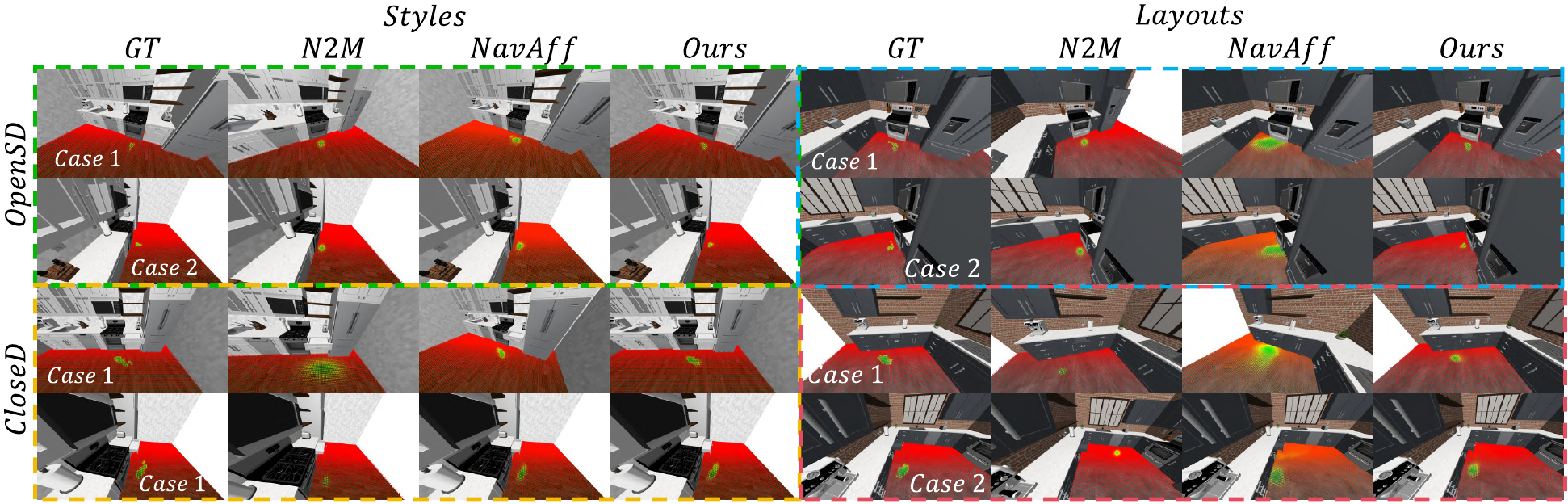}
 % \vspace{-0.3cm}
 \caption{Illustration of qualitative results for predicting FloAff using different methods on the CloseD and OpenSD tasks (FloAff-Kitchen benchmarks). Under different styles and layouts, each task is shown in two cases. Please see the supplementary material for more visualizations.}
 \label{fig4}
 \vspace{-0.2cm}
\end{figure*}

\noindent\textbf{Implementation Details.}

Unless otherwise specified, the cropping radius in Eq.~(\ref{eq:crop}) is set to $r=1.3$ m for the \textit{Styles} benchmark and $r=1.2$ m for the \textit{Layouts} benchmark. AxisNet is pre-trained for target orientation estimation using local point clouds with the Adam optimizer \cite{kingma2014adam}. During PFAL, the best checkpoint from the previous stage initializes the subsequent stage, while the pre-trained AxisNet remains frozen. We further apply random yaw perturbations within $\pm5^\circ$ for data augmentation. The loss weights are set to $\lambda=1.0$ and $\beta=1.0$. Each progressive stage is trained for $30$ epochs using Adam with a learning rate of $2\times10^{-4}$ and a batch size of $8$. All experiments are conducted on three NVIDIA RTX 3090 GPUs.

\subsection{Comparative Studies}

\noindent\textbf{Comparative studies on FloAff prediction.}
Table~\ref{table1} compares our method with state-of-the-art approaches on three benchmarks. Our method consistently achieves the best performance across all evaluation metrics, indicating its effectiveness in predicting manipulation-oriented FloAff from egocentric observations. On the MoMa-Kitchen benchmark, our method substantially outperforms all baselines, reducing RMSE from $0.147$ to $0.073$ and improving PCC from $0.680$ to $0.796$ over the previous state of the art, NavAff. These improvements suggest that CFAR helps alleviate representation ambiguity by focusing the model on affordance-relevant interaction geometry rather than redundant scene context and arbitrary orientations. Meanwhile, PFAL further enhances generalization by progressively transferring shared FloAff priors across heterogeneous manipulation skills.

The advantages become more evident on the proposed FloAff-Kitchen benchmarks, which introduce challenging distribution variations. Under the \emph{Styles} benchmark, our method maintains high prediction accuracy despite significant appearance variations, indicating that the learned representation is less sensitive to texture and furniture style changes. Compared with \emph{Styles}, the \emph{Layouts} benchmark is more challenging because unseen layouts alter interaction geometry rather than only visual appearance, requiring stronger structural generalization. Nevertheless, our method consistently outperforms existing approaches, suggesting that CFAR and PFAL provide robust FloAff representations under both appearance and structural variations.

\renewcommand\arraystretch{0.7}
\begin{table}[t!] \small
\centering
\setlength{\tabcolsep}{2pt}
\begin{tabular}{lcccc}
\toprule
\textbf{Method{(Venue)}} & \textbf{RMSE} $\downarrow$ & \textbf{logMSE} $\downarrow$ & \textbf{PCC} $\uparrow$ & \textbf{SIM} $\uparrow$ \\ 
\midrule
\textbf{MoMa-Kitchen} &  &  &  &  \\
PointNet++{(NeurIPS'17)} & 0.164 & 0.0142 & 0.565 & 0.589 \\
VoteNet{(ICCV'19)} & 0.167 & 0.0143 & 0.543 & 0.570 \\
H3DNet{(ECCV'20)} & 0.174 & 0.0156 & 0.503 & 0.522 \\
NavAff{(ICCV'25)} & 0.147 & 0.0115 & 0.680 & 0.696 \\ 
% \textbf{Ours(Rigid->Articulated)} & 0.091 & 0.0043 & 0.682 & 0.686 \\
\textbf{Ours} & \textbf{0.073} & \textbf{0.0028} & \textbf{0.796} & \textbf{0.794} \\
\midrule
\textbf{FloAff-Kitchen(\emph{Styles})} &  &  &  &  \\
NavAff{(ICCV'25)} & 0.064 & 0.0025 & 0.395 & 0.337 \\ 
\textbf{Ours} & \textbf{0.029} & \textbf{0.0004} & \textbf{0.863} & \textbf{0.862} \\
\midrule
\textbf{FloAff-Kitchen(\emph{Layouts})} &  &  &  &  \\
NavAff{(ICCV'25)} & 0.080 & 0.0041 & 0.238 & 0.225 \\ 
\textbf{Ours} & \textbf{0.035} & \textbf{0.0006} & \textbf{0.738} & \textbf{0.721} \\
\bottomrule
\end{tabular}
\caption{Quantitative evaluation of FloAff prediction across different datasets.}
\label{table1}
\vspace{-0.25cm}
\end{table}

\renewcommand\arraystretch{0.7}
\begin{table*}[t!]
\centering
\small
\setlength{\tabcolsep}{8pt}
\begin{tabular}{lccccccc}
\toprule
\multirow{2.5}{*}{\textbf{Method{(Venue)}}} & \multicolumn{3}{c}{\textbf{\emph{Styles}}} & \multicolumn{3}{c}{\textbf{\emph{Layouts}}} \\
\cmidrule(lr){2-4} \cmidrule(lr){5-7}
& $SR_{\text{front}}$ & $SR_{\text{side}}$ & $SR_{\text{all}}$ & $SR_{\text{front}}$ & $SR_{\text{side}}$ & $SR_{\text{all}}$ \\
\midrule
NavAff{(ICCV'25)}        & 0.28 $\pm$ {0.05} & 0.46 $\pm$ {0.04} & 0.37 $\pm$ {0.03} & 0.28 $\pm$ {0.04} & 0.26 $\pm$ {0.03} & 0.27 $\pm$ {0.03} \\
C2F-Exp{(AAAI'26)}       & 0.53 $\pm$ {0.05} & 0.48 $\pm$ {0.04} & 0.50 $\pm$ {0.03} & 0.50 $\pm$ {0.05} & 0.47 $\pm$ {0.04} & 0.49 $\pm$ {0.05} \\
N2M{(ICML'26)}           & 0.56 $\pm$ {0.04} & 0.33 $\pm$ {0.04} & 0.44 $\pm$ {0.03} & 0.72 $\pm$ {0.03} & 0.22 $\pm$ {0.03} & 0.47 $\pm$ {0.02} \\
\textbf{Ours}                 & \textbf{0.85 $\pm$ {0.04}} & \textbf{0.85 $\pm$ {0.03}} & \textbf{0.85 $\pm$ {0.01}} & \textbf{0.81 $\pm$ {0.03}} & \textbf{0.77 $\pm$ {0.03}} & \textbf{0.79 $\pm$ {0.02}} \\
\midrule
Oracle                        & 0.87 $\pm$ {0.04} & 0.88 $\pm$ {0.03} & 0.88 $\pm$ {0.03} & 0.86 $\pm$ {0.02} & 0.87 $\pm$ {0.03} & 0.86 $\pm$ {0.02}  \\
\bottomrule
\end{tabular}
\caption{Quantitative evaluation of FloAff-grounded MoMa performance on the FloAff-Kitchen benchmarks (multi-task). Oracle describes the MoMa performance at the optimal pose.}
\label{table2}
\vspace{-0.3cm}
\end{table*}

\begin{table}[t]
\centering
\small 
\setlength{\tabcolsep}{3pt}
\begin{tabular}{lcccc}
\toprule
\textbf{Method} & \textbf{PnPC2C} & \textbf{OpenSD} & \textbf{CloseD} & \textbf{CloseDD} \\
\midrule
% \rowcolor[HTML]{E8E8FF}
\multicolumn{5}{l}{\textbf{\emph{Styles}}} \\
NavAff    & 0.61 $\pm$ {0.04} & 0.50 $\pm$ {0.04} & 0.97 $\pm$ {0.01} & 0.57 $\pm$ {0.04} \\
C2F-Exp   & 0.44 $\pm$ {0.04} & 0.30 $\pm$ {0.04} & 0.72 $\pm$ {0.04} & 0.54 $\pm$ {0.04} \\
N2M       & 0.70 $\pm$ {0.03} & 0.48 $\pm$ {0.01} & 0.94 $\pm$ {0.01} & 0.58 $\pm$ {0.03} \\
\textbf{Ours} & \textbf{0.83} $\pm$ {\textbf{0.03}} & \textbf{0.86} $\pm$ {\textbf{0.03}} & \textbf{0.98} $\pm$ {\textbf{0.02}} & \textbf{0.72} $\pm$ {\textbf{0.03}} \\
% \midrule
% Oracle & 0.82 $\pm$ {0.03} & 0.88 $\pm$ {0.02} & 0.98 $\pm$ {0.02} & 0.86 $\pm$ {0.03} \\
\midrule
% \rowcolor[HTML]{E8E8FF}
\multicolumn{5}{l}{\textbf{\emph{Layouts}}} \\
NavAff    & 0.34 $\pm$ {0.05} & 0.27 $\pm$ {0.03} & 0.05 $\pm$ {0.01} & 0.44 $\pm$ {0.03} \\
C2F-Exp   & 0.41 $\pm$ {0.04} & 0.41 $\pm$ {0.04} & 0.75 $\pm$ {0.04} & 0.57 $\pm$ {0.04} \\
N2M       & 0.28 $\pm$ {0.03} & 0.78 $\pm$ {0.02} & 0.31 $\pm$ {0.02} & 0.42 $\pm$ {0.01} \\
\textbf{Ours} & \textbf{0.56} $\pm$ {\textbf{0.03}} & \textbf{0.79} $\pm$ {\textbf{0.03}} & \textbf{0.99} $\pm$ {\textbf{0.04}} & \textbf{0.70} $\pm$ {\textbf{0.03}} \\
% \midrule
% Oracle & 0.66 $\pm$ {0.03} & 0.88 $\pm$ {0.03} & 0.99 $\pm$ {0.01} & 0.91 $\pm$ {0.04} \\
\bottomrule
\end{tabular}
\caption{Quantitative evaluation of FloAff-grounded MoMa performance on the FloAff-Kitchen benchmarks. (single-task \& all-view).}
\label{table3}
\vspace{-0.3cm}
\end{table}

\noindent\textbf{Comparative studies on FloAff-grounded multi-task MoMa.}
Table~\ref{table2} reports the downstream MoMa performance when the predicted FloAff maps are used for robot base placement. Our method consistently achieves the highest success rates across all benchmark settings, indicating that the learned FloAff representations effectively translate into improved manipulation performance beyond affordance prediction accuracy. Compared with existing methods, our approach substantially improves the overall success rate on both \emph{Styles} and \emph{Layouts}. The improvement is particularly notable on \emph{Layouts}, where unseen kitchen layouts introduce structural variations in interaction geometry rather than only appearance changes. Despite this challenging setting, our method achieves a 79\% overall success rate, significantly outperforming all baselines and approaching the Oracle policy. These results suggest that CFAR provides more consistent interaction geometry representations, while PFAL facilitates transferable FloAff priors across heterogeneous manipulation skills. The qualitative results in Figure \ref{fig4} further show that our predicted FloAff maps better match the GT distributions in terms of location, shape, and spatial extent.

The results further demonstrate robustness to viewpoint variations. Under \emph{Styles}, both front-view and side-view observations achieve an 85\% success rate, indicating that the canonical representation reduces the impact of viewpoint-induced representation variations. Even under the more challenging \emph{Layouts} setting, the performance gap between front-view and side-view remains relatively small (81\% vs.\ 77\%), while existing methods achieve consistently lower performance under both viewpoints. Overall, these results indicate that CFAR and PFAL together provide robust FloAff prediction under appearance, structural, and viewpoint variations, enabling downstream MoMa performance close to the Oracle upper bound.

\noindent\textbf{Comparative studies on FloAff-grounded single-task MoMa.}
To further analyze the contribution of CFAR independent of PFAL, we evaluate FloAff-grounded MoMa under single-task training. As shown in Table~\ref{table3}, the proposed CFAR consistently achieves the best performance across all manipulation tasks under both \emph{Styles} and \emph{Layouts}, indicating that canonical representation learning provides a strong basis for robust FloAff prediction and downstream manipulation.

The improvements are particularly evident for articulated-object manipulation, including \emph{OpenSD} and \emph{CloseDD}. These tasks are highly sensitive to target orientation because feasible robot base placements depend on hinge-related interaction geometry. By transforming target-centered observations into a unified coordinate system, CFAR reduces orientation-induced representation variations and enables more reliable FloAff prediction compared with existing methods. For \emph{CloseD}, our method achieves near-saturated performance under both benchmark settings, suggesting that the learned canonical representation effectively captures the local geometry required for manipulation.

Although \emph{Layouts} introduces substantially larger structural variations than \emph{Styles}, CFAR consistently improves performance across all four tasks. This result indicates that canonical interaction geometry generalizes beyond appearance variations to unseen interaction configurations. Overall, these experiments highlight the importance of learning canonical representations for robust FloAff prediction and downstream MoMa.

\subsection{Ablation Studies}

Tables~\ref{table4} and~\ref{table5} present ablation studies on MoMa-Kitchen and the more challenging \emph{Layouts} benchmark, respectively. Removing any component consistently degrades performance, indicating that each proposed module contributes to FloAff prediction. On MoMa-Kitchen, removing SDL causes the largest performance drop, suggesting that region-level supervision is important for recovering complete feasible robot base placement regions beyond point-wise regression. Removing the Feature Distillation (FD) loss also degrades all metrics, indicating that preserving shared FloAff priors benefits progressive adaptation across heterogeneous manipulation skills. In contrast, removing point cloud cropping results in only a slight degradation, suggesting that the benefit of eliminating redundant spatial context is less evident in the relatively constrained MoMa-Kitchen scenes.

The trends become more evident on the \emph{Layouts} benchmark, where scene geometry varies substantially across kitchens. Removing either target-centered cropping or orientation alignment significantly reduces PCC and SIM, indicating that both components of CFAR are important for learning consistent interaction geometry under layout variations. Meanwhile, removing FD still degrades performance, suggesting that preserving transferable FloAff priors remains beneficial under large structural scene variations. Overall, these results highlight the complementary roles of CFAR and PFAL: CFAR learns compact canonical representations by reducing irrelevant spatial variations, while PFAL improves transferability across heterogeneous manipulation skills.

\emph{Please refer to the supplementary material for further parametric studies and more visualizations.}

\renewcommand\arraystretch{0.7}
\begin{table}[t!] \small
\centering
\setlength{\tabcolsep}{6pt}
\begin{tabular}{lcccc}
\toprule
\textbf{Ablation} & \textbf{RMSE} $\downarrow$ & \textbf{logMSE} $\downarrow$ & \textbf{PCC} $\uparrow$ & \textbf{SIM} $\uparrow$ \\ 
\midrule
w/o Crop & 0.075 & 0.0029 & 0.789 & 0.791 \\
w/o SDL & 0.089 & 0.0041 & 0.707 & 0.708 \\
w/o FD & 0.088 & 0.0041 & 0.729 & 0.729 \\

\midrule
\textbf{Ours(Full)} & \textbf{0.073} & \textbf{0.0028} & \textbf{0.796} & \textbf{0.794} \\
\bottomrule
\end{tabular}
\caption{Ablation studies of CFAR and PFAL on the MoMa-Kitchen benchmark.}
\label{table4}
% \vspace{-0.1cm}
\end{table}

\renewcommand\arraystretch{0.7}
\begin{table}[t!] \small
\centering
\setlength{\tabcolsep}{5pt}
\begin{tabular}{lcccc}
\toprule
 & \multicolumn{4}{c}{\textbf{\emph{Layouts}}} \\
\cmidrule{2-5}
\textbf{Ablation} & \textbf{RMSE} $\downarrow$ & \textbf{logMSE} $\downarrow$ & \textbf{PCC} $\uparrow$ & \textbf{SIM} $\uparrow$ \\ 
\midrule
w/o Crop   & 0.049 & 0.0014 & 0.552 & 0.541 \\
w/o SDL   & 0.051 & 0.0014 & 0.638 & 0.618 \\
w/o Align & 0.051 & 0.0013 & 0.683 & 0.669 \\
w/o FD   & 0.041 & 0.0010 & 0.719 & 0.703 \\
\midrule
\textbf{Ours(Full)} & \textbf{0.035} & \textbf{0.0006} & \textbf{0.738} & \textbf{0.721} \\
\bottomrule
\end{tabular}
\caption{Ablation studies of CFAR and PFAL on the FloAff-Kitchen (Layouts) benchmark.}
\label{table5}
\vspace{-0.3cm}
\end{table}

% \subsection{Parametric Studies}

\section{Conclusion}

This paper presented a unified framework for FloAff prediction in MoMa tasks. We identified two key challenges in existing methods: representation ambiguity caused by irrelevant spatial variations and limited transferability of affordance learning across heterogeneous manipulation skills. To address these challenges, we proposed CFAR to learn canonical interaction geometry through target-conditioned representation transformation, and PFAL to progressively acquire transferable FloAff priors through progressive adaptation. We further established the FloAff-Kitchen benchmark for systematic evaluation under viewpoint, style, and layout variations. Extensive experiments on MoMa-Kitchen and FloAff-Kitchen demonstrated consistent improvements in both FloAff prediction and downstream MoMa performance, highlighting the effectiveness and generalization capability of the proposed framework. Future work will extend the proposed method to more diverse manipulation skills and real-world MoMa scenarios.

\bibliography{aaai2027}

\clearpage
\appendix

\twocolumn[
\begin{center}
    {\LARGE\bfseries
    Technical Supplement for ``FloAff-Kitchen: Bridging Navigation and Manipulation via Canonical and Progressive Floor Affordance Learning''\par}
    \vspace{1.5em}
\end{center}
]

\section{Implementation Details of AxisNet}

AxisNet is a lightweight orientation predictor that estimates the target yaw angle $\theta_o$ for canonical orientation alignment in CFAR. It takes the cropped egocentric point cloud and the target-object feature as inputs. Specifically, each input point is represented as
\begin{equation}
    \mathbf{x}_i=[\bar{\mathbf{p}}_i;m_i]\in\mathbb{R}^{4},
\end{equation}
where $\bar{\mathbf{p}}_i$ denotes the normalized 3D coordinate relative to the target center and $m_i$ denotes the corresponding target-object feature.

AxisNet adopts a lightweight PointNet-style architecture, which extracts point-wise features through shared convolutional layers and aggregates them into a global scene representation via max pooling. To avoid the discontinuity of direct angular regression, AxisNet predicts an $\ell_2$-normalized orientation vector:
\begin{equation}
    \hat{\mathbf{q}}_o
    =[\sin(\hat{\theta}_o),\cos(\hat{\theta}_o)].
\end{equation}
Given the ground-truth orientation vector $\mathbf{q}_o$, AxisNet is optimized by minimizing the orientation prediction loss:
\begin{equation}
    \mathcal{L}_{\mathrm{axis}}
    =\operatorname{MSE}(\hat{\mathbf{q}}_o,\mathbf{q}_o).
\end{equation}

During training, we apply random yaw perturbations uniformly sampled within $\pm90^\circ$ to each input point cloud and update its orientation label accordingly. Gaussian coordinate jitter with a standard deviation of $0.005$ is further introduced for data augmentation. AxisNet is trained for $200$ epochs using the Adam optimizer with a learning rate of $1\times10^{-4}$ and a batch size of $32$. The checkpoint with the lowest validation loss is selected, and the resulting AxisNet is kept frozen during PFAL training.

\section{Parametric Studies}

The effect of the crop radius is evaluated on both the FloAff-Kitchen \textit{Layouts} and \textit{Styles} benchmarks. Specifically, Cosine Similarity (SIM) measures the spatial agreement between the predicted and ground-truth affordance maps, reflecting their consistency in terms of location, shape, and extent. This property is particularly relevant to downstream MoMa, where inaccurate affordance regions may lead to suboptimal robot base placements even when the regression error remains relatively low.

As shown in Table~\ref{tab:crop_size_styles}, a crop radius of $1.30\,\mathrm{m}$ achieves the highest PCC and SIM on \textit{Styles}, suggesting the strongest agreement between the predicted and ground-truth affordance distributions. Since this benchmark preserves the underlying kitchen layouts while varying furniture styles, textures, and object appearances, the interaction geometry around the manipulation target remains relatively consistent. These results suggest that a radius of $1.30\,\mathrm{m}$ preserves sufficient affordance-relevant geometry while reducing the influence of appearance-related variations. Considering that SIM directly evaluates the spatial consistency of manipulation-effective regions, we set the crop radius to $1.30\,\mathrm{m}$ for the \textit{Styles} benchmark.

Table~\ref{tab:crop_size_layouts} reports the results on \textit{Layouts}, where structural variations across kitchen scenes introduce stronger challenges for local geometry modeling. A crop radius of $1.20\,\mathrm{m}$ achieves the best overall performance, indicating a better trade-off between preserving task-relevant interaction geometry and suppressing irrelevant spatial context. A smaller radius may exclude geometric information related to reachability and collision-free manipulation, whereas a larger radius introduces additional scene variations that are less relevant to FloAff prediction. Therefore, the crop radius is set to $1.20\,\mathrm{m}$ for the \textit{Layouts} benchmark.

\begin{table}[t!]
\small
\centering
\setlength{\tabcolsep}{7pt}
\begin{tabular}{lcccc}
\toprule
 & \multicolumn{4}{c}{\textbf{\emph{Styles}}} \\
\cmidrule{2-5}
\textbf{Crop (m)}
& \textbf{RMSE} $\downarrow$
& \textbf{logMSE} $\downarrow$
& \textbf{PCC} $\uparrow$
& \textbf{SIM} $\uparrow$ \\
\midrule
1.10 & 0.028 & 0.0004 & 0.848 & 0.848 \\
1.15 & 0.026 & 0.0004 & 0.848 & 0.848 \\
1.20 & 0.027 & 0.0004 & 0.854 & 0.855 \\
1.25 & 0.026 & 0.0004 & 0.846 & 0.846 \\
1.30 & 0.029 & 0.0004 & \textbf{0.863} & \textbf{0.862} \\
1.35 & 0.027 & 0.0004 & 0.855 & 0.856 \\
1.40 & \textbf{0.025} & \textbf{0.0003} & 0.854 & 0.854 \\
\bottomrule
\end{tabular}
\caption{Parametric study on the crop radius using the FloAff-Kitchen (Styles) benchmark.}
\label{tab:crop_size_styles}
\end{table}

\begin{table}[t!]
\small
\centering
\setlength{\tabcolsep}{7pt}
\begin{tabular}{lcccc}
\toprule
 & \multicolumn{4}{c}{\textbf{\emph{Layouts}}} \\
\cmidrule{2-5}
\textbf{Crop (m)} 
& \textbf{RMSE} $\downarrow$ 
& \textbf{logMSE} $\downarrow$ 
& \textbf{PCC} $\uparrow$ 
& \textbf{SIM} $\uparrow$ \\
\midrule
1.10 & 0.050 & 0.0014 & 0.662 & 0.647 \\
1.15 & 0.046 & 0.0011 & 0.729 & 0.712 \\
1.20 & \textbf{0.035} & \textbf{0.0006} & \textbf{0.738} & \textbf{0.721} \\
1.25 & 0.037 & 0.0007 & 0.732 & 0.715 \\
1.30 & 0.041 & 0.0009 & 0.725 & 0.709 \\
1.35 & 0.043 & 0.0009 & 0.720 & 0.704 \\
1.40 & 0.036 & 0.0007 & 0.716 & 0.700 \\
\bottomrule
\end{tabular}
\caption{Parametric study on the crop radius using the FloAff-Kitchen (Layouts) benchmark.}
\label{tab:crop_size_layouts}
\end{table}

\section{Cross-scene Floor Affordance Benchmark}

This section provides detailed descriptions of the proposed FloAff-Kitchen benchmark, including scene generation, candidate robot base pose generation, rollout-based affordance annotation, multi-view observation collection, and dataset statistics.

%%%%%%%%%%%%%%%%%%%%%%%%%%%%%%%%%%%%%%%%%%%%%%%%%%%%%%%%%%%%%%%%%%%%%%%%%%%%%%%
\subsection{Scene Generation}

The benchmark is constructed in RoboCasa \cite{robocasa2024}, which provides procedurally configurable kitchen environments. Each scene is specified by a layout identifier and a furniture style identifier, which jointly determine the spatial arrangement of kitchen components, articulated objects, object instances, and visual appearances. By combining different layouts and styles, we obtain diverse kitchen scenes with variations in cabinet configurations, drawer placements, countertop arrangements, appliance locations, and appearances.

Following the benchmark setting of MoMa-Kitchen \cite{zhang2025moma}, we consider four representative MoMa tasks:

\begin{itemize}
    \item \textbf{PnPCounterToCab (PnPC2C):} pick an object from the countertop and place it into a cabinet;
    \item \textbf{OpenSingleDoor (OpenSD):} open a microwave door;
    \item \textbf{CloseDrawer (CloseD):} close an opened drawer;
    \item \textbf{CloseDoubleDoor (CloseDD):} close a double-door cabinet.
\end{itemize}

These tasks cover both object rearrangement and articulated manipulation, requiring robots to reason about reachability, collision avoidance, target visibility, and task-specific interaction constraints. To evaluate scene generalization, we construct two benchmark settings:

\begin{itemize}
    \item \textbf{Styles Benchmark:} preserves kitchen layouts while varying furniture styles, textures, and object appearances.
    \item \textbf{Layouts Benchmark:} varies kitchen layouts while preserving the manipulation task definitions.
\end{itemize}

All train/test splits are performed at the scene level.

%%%%%%%%%%%%%%%%%%%%%%%%%%%%%%%%%%%%%%%%%%%%%%%%%%%%%%%%%%%%%%%%%%%%%%%%%%%%%%%
\subsection{Candidate Robot Base Pose Generation}

For each instantiated kitchen scene, we generate candidate robot base poses that are potentially feasible for downstream manipulation. Specifically, RGB-D observations captured from multiple fixed cameras are fused into a global point cloud
$\mathcal P=\{\mathbf p_i\}_{i=1}^{N}$,
where each point
$\mathbf p_i=(x_i,y_i,z_i)$
is represented in the world coordinate system. Ground points are extracted according to their height and discretized into a two-dimensional occupancy grid with a resolution of $0.05$m.

Instead of exhaustively searching the entire floor space, candidate poses are sampled within a target-centered region. Let
$\mathbf c=(x_c,y_c)$
denote the projection of the manipulation target onto the floor plane and
$\theta_o$
denote its facing orientation. A ground point is retained if it satisfies:
\begin{equation} 
|x_i^{o}| \le r_x,\qquad
|y_i^{o}| \le r_y,
\label{eq:supp_candidate}
\end{equation}
where
$(x_i^{o},y_i^{o})$
are the point coordinates expressed in the target coordinate frame, and
$r_x,r_y$
are task-dependent search ranges. Each remaining grid cell defines a candidate robot base pose
$\mathbf b_i=(x_i,y_i,\theta_i)$,
where the orientation is initialized toward the manipulation target.

%%%%%%%%%%%%%%%%%%%%%%%%%%%%%%%%%%%%%%%%%%%%%%%%%%%%%%%%%%%%%%%%%%%%%%%%%%%%%%%
\subsection{Collision-aware Filtering}

To remove physically infeasible poses before manipulation evaluation, candidate poses are filtered according to collision and reachability constraints. The robot body is approximated by three collision regions corresponding to the mobile base, torso, and manipulator workspace. Let
$\mathcal C_{\rm base},
\mathcal C_{\rm torso},
\mathcal C_{\rm arm}$
denote the corresponding collision indicators. A candidate pose is retained only if
$\mathcal C_{\rm base}=0$,
$\mathcal C_{\rm torso}=0$,
and
$\mathcal C_{\rm arm}=0$.

A reachability constraint is further imposed:
$\|
\mathbf b_{i,xy}
-
\mathbf c
\|_2
<
r_{\rm arm},$
where
$r_{\rm arm}$
denotes the maximum reachable manipulation radius. The remaining candidate poses form the rollout set
$\mathcal B
=
\{
\mathbf b_i
\}_{i=1}^{M}$,
where $M$ is the number of candidate poses after filtering.

%%%%%%%%%%%%%%%%%%%%%%%%%%%%%%%%%%%%%%%%%%%%%%%%%%%%%%%%%%%%%%%%%%%%%%%%%%%%%%%

\begin{table}[t]
\centering
\begin{tabular}{lccccc}
\toprule
\multirow{2}{*}{\textbf{Task}} & \multicolumn{2}{c}{\textbf{Train}} & \multicolumn{2}{c}{\textbf{Test}} & \multirow{2}{*}{\textbf{Total}} \\
\cmidrule(lr){2-3} \cmidrule(lr){4-5}
& \textbf{Front} & \textbf{Side} & \textbf{Front} & \textbf{Side} & \\
\midrule
\textbf{Styles} &  &  &  &  &  \\
PnPC2C  & 960  & 1920 & 200 & 400 & 3480 \\
OpenSD  & 1000 & 2000 & 200 & 400 & 3600 \\
CloseD  & 1000 & 1780 & 200 & 400 & 3380 \\
CloseDD & 1000 & 1630 & 200 & 320 & 3150 \\
\textbf{Total} & 3960 & 7330 & 800 & 1520 & 13610 \\
\midrule
\textbf{Layouts} &  &  &  &  &  \\
PnPC2C  & 1000 & 1200 & 200 & 200 & 2600 \\
OpenSD  & 1000 & 1400 & 200 & 200 & 2800 \\
CloseD  & 1000 & 1783 & 200 & 199 & 3182 \\
CloseDD & 1000 & 1190 & 200 & 200 & 2590 \\
\textbf{Total} & 4000 & 5573 & 800 & 799 & 11172 \\
\bottomrule
\end{tabular}
\caption{Statistics of RoboCasa dataset.}
\label{tab:dataset_statistics}
\end{table}

\begin{figure*}[!t]
 \centering
 \includegraphics[width=1.0\linewidth]{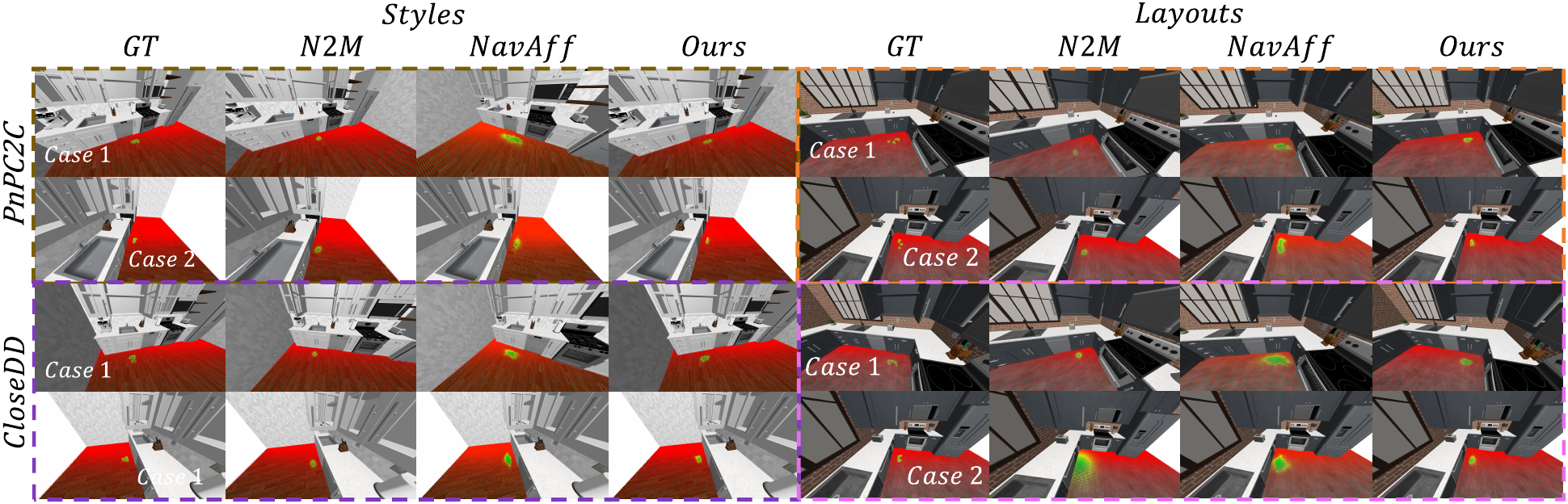}
 \caption{Illustration of qualitative results for predicting FloAff using different methods on the PnPC2C and CloseDD tasks (FloAff-Kitchen benchmarks). Under different styles and layouts, each task is shown in two cases.}
 \label{fig6}
\end{figure*}

\begin{figure}[!t]
 \centering
 \includegraphics[width=1.0\linewidth]{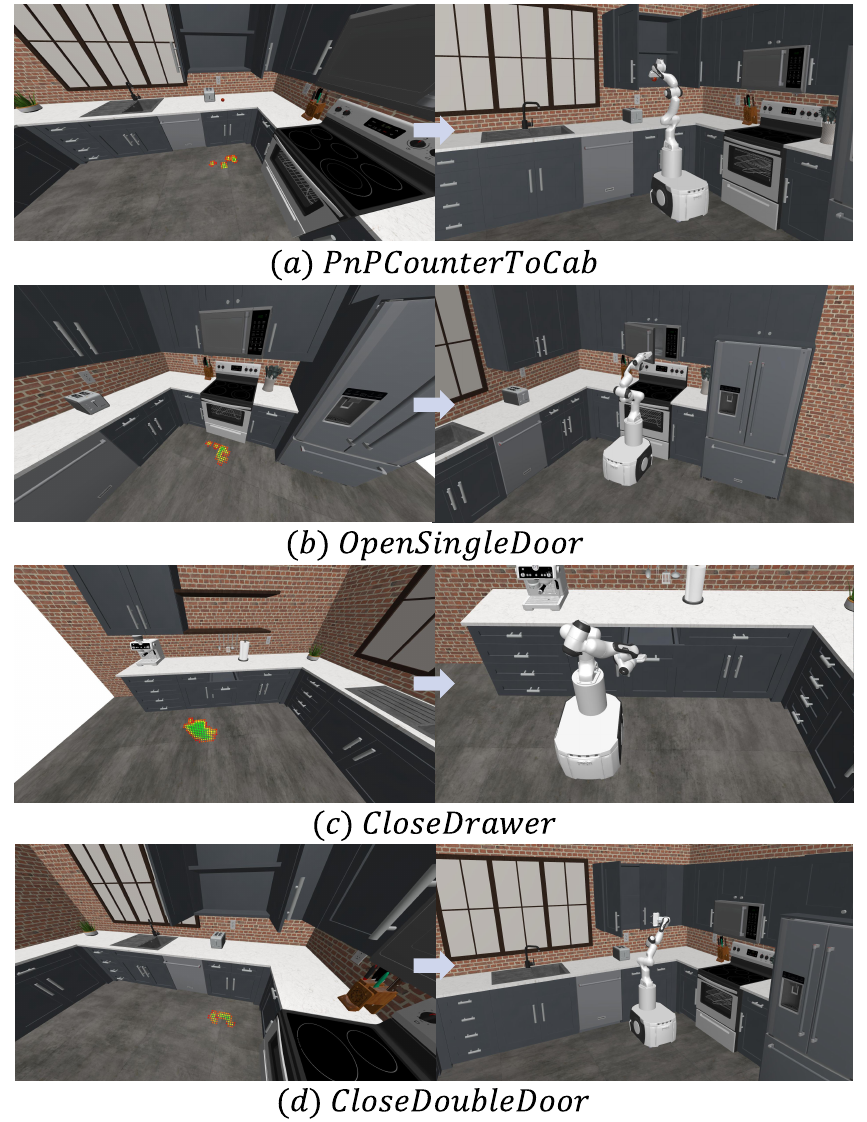}
 \caption{Illustration of FloAff predictions and the four downstream MoMa tasks completed based on them.}
 \label{fig7}
\end{figure}

\subsection{Rollout-based Affordance Annotation}

For each feasible candidate pose, we evaluate whether it supports successful manipulation using a pretrained task-specific manipulation policy. Given a candidate pose
$\mathbf b_i$,
the robot is initialized at the corresponding location and executes the policy
$\pi_\tau$
for at most
$H$
simulation steps, where
$\tau$
denotes the manipulation task. The rollout terminates once the task succeeds. Each candidate pose is assigned a binary affordance label:
\begin{equation}
y_i
=
\mathbf{1}
\left[
\mathrm{Success}
(
\pi_\tau,
\mathbf b_i
)
\right],
\label{eq:supp_rollout}
\end{equation}
where
$y_i=1$
indicates successful manipulation from pose
$\mathbf b_i$.

The rollout labels are projected onto the floor plane to construct the affordance annotation:
\begin{equation}
\mathcal Y
=
\left\{
(\mathbf b_i,y_i)
\right\}_{i=1}^{M},
\label{eq:supp_affmap}
\end{equation}
where
$M$
denotes the number of evaluated candidate poses. These rollout-generated labels directly characterize whether a robot base placement can support downstream manipulation under the corresponding task dynamics.

%%%%%%%%%%%%%%%%%%%%%%%%%%%%%%%%%%%%%%%%%%%%%%%%%%%%%%%%%%%%%%%%%%%%%%%%%%%%%%%
\subsection{Multi-View Observation Collection}

After rollout annotation, egocentric RGB-D observations are collected from multiple viewpoints around the manipulation target. Let
$(x^o,y^o)$
denote the camera position expressed in the target coordinate frame. Front-view observations are sampled from:
\begin{equation}
\mathcal V_{\rm front}
=
\{
(x^o,y^o)
~
|
~
|x^o|
<
r_f^x,
~
y^o
<
-r_f^y
\},
\end{equation}
whereas side-view observations are sampled from:
\begin{equation}
\mathcal V_{\rm side}
=
\{
(x^o,y^o)
~
|
~
|x^o|
>
r_s^x,
~
|y^o|
<
r_s^y
\}.
\end{equation}

For each viewpoint, the robot heading is aligned toward the manipulation target to maintain target visibility. Compared with front-view observations, side-view observations introduce stronger occlusion and viewpoint ambiguity, particularly for articulated manipulation tasks. Both viewpoints share identical rollout-generated affordance labels, enabling controlled evaluation of viewpoint robustness.

%%%%%%%%%%%%%%%%%%%%%%%%%%%%%%%%%%%%%%%%%%%%%%%%%%%%%%%%%%%%%%%%%%%%%%%%%%%%%%%
\subsection{Dataset Statistics}

After removing scenes without positive affordance annotations, the benchmark contains 24,782 RGB-D observations paired with rollout-generated FloAff labels. Among them, 13,610 samples belong to the \emph{Styles} benchmark and 11,172 samples belong to the \emph{Layouts} benchmark. All train/test splits are performed at the scene level, using five training scenes and one unseen testing scene for each manipulation task. Detailed statistics are summarized in Table~\ref{tab:dataset_statistics}.

\section{More Visualizations}

To supplement the visualizations in the paper, Figure \ref{fig6} illustrates the qualitative results for predicting FloAff using different methods on the PnPC2C and CloseDD tasks (FloAff-Kitchen benchmarks). Under different styles and layouts, each task is shown in two cases. Figure \ref{fig7} shows examples of four downstream MoMa tasks completed using our FloAff predictions.

\end{document}